\documentclass[11pt]{article}
\usepackage[utf8]{inputenc}
\usepackage{textgreek}

\usepackage[final]{acl}

\usepackage{times}
\usepackage{latexsym}
\usepackage{amsfonts}
\usepackage[T1]{fontenc}
\usepackage[utf8]{inputenc}
\usepackage{amsmath}
\usepackage{microtype}
\usepackage{inconsolata}
\usepackage{graphicx}
\usepackage{booktabs}
\usepackage{enumitem}
\usepackage{float}
\usepackage{placeins}
\usepackage{array}
\usepackage{titlesec}
\usepackage{balance}
\usepackage{needspace}

\newcommand{\sig}{^{\dag}}
\newcommand{\nsig}{^{\mathrm{ns}}}

\title{Making COMET Comparable Across Scripts:\\
       Diagnosis and Correction of Tokeniser-Induced\\
       Script Bias in Indic MT Evaluation}

\author{G.~L. John Salvin \quad Swapnil Hingmire \\
  Mehta Family School of Data Science and Artificial Intelligence \\
  Indian Institute of Technology Palakkad, Kerala, India \\
  \texttt{142503002@smail.iitpkd.ac.in} \quad
  \texttt{swapnilh@iitpkd.ac.in}}

\begin{document}
\maketitle
\begin{abstract}
COMET reports translation quality as a single number, and that number is
routinely compared across target languages written in different scripts.
Such a comparison assumes \emph{Script Invariance}: the score should not
depend on the writing system that carries the target. We test it on
IndicMT~Eval by re-encoding the target into Latin script, which changes
orthographic form while holding content and human ratings fixed. Script
identity then accounts for $22.9\%$ of native-script COMET variance, and
agreement with annotators falls in all five languages studied. We trace
the effect to the tokeniser and measure it with three label-free
diagnostics. The bias is two faults, not one. Scores from different
scripts occupy incompatible ranges, and within a single script the metric
orders translations less accurately. No order-preserving transform of the
score can repair the second fault. The first is removed exactly by
COMET-QN, which maps the score distribution of each (language, script)
pair onto a shared reference. Pooled agreement with annotators rises from
$0.300$ to $0.399$, which is what makes scores from different scripts
safe to place on one axis, and every within-language ordering is provably
preserved. A regressor over parity features recovers a further $17.1\%$
of the lost sensitivity. The remainder belongs to the encoder, and no
post-processing can reach it. We therefore recommend publishing the
normalised score, the three diagnostics, and the identity of the
tokeniser they were computed against, so that a reader can tell how much
of a score reflects translation quality and how much reflects the writing
system.\footnote{Data and code:
\url{https://github.com/John-salvin/script-bias-comet-normalisation}.}
\end{abstract}

\section{Introduction}\label{sec:intro}
Learned metrics have displaced surface metrics such as
BLEU~\citep{DBLP:conf/acl/PapineniRWZ02} as the primary quality signal in
machine translation. COMET~\citep{DBLP:conf/emnlp/ReiSFL20} projects
source, hypothesis and reference into one multilingual space and scores
adequacy by proximity within it. On high-resource European pairs its
agreement with Multidimensional Quality Metrics
(MQM;~\citealp{lommel-etal-2014-mqm}) annotation far exceeds that of
$n$-gram metrics~\citep{DBLP:journals/tacl/FreitagBBEGMNRR21}, and it now
underpins system rankings at WMT and throughout the field.

Emitting a single scalar invites that scalar to travel: between systems,
between test sets, and, more consequentially, between target languages
written in different scripts. That last comparison carries an unstated
premise. Call it \emph{Script Invariance} (SI). Writing $\pi_p(\cdot)$
for a transliteration that re-encodes a string into script $p$ without
disturbing its linguistic content, SI asserts that for every admissible
target script $p$,
\begin{equation}
\mathrm{COMET}(s,\, \pi_p(t),\, \pi_p(r)) = \mathrm{COMET}(s,\,t,\,r).
\label{eq:si}
\end{equation}
The premise is operational rather than architectural. No component of
COMET consults script identity; it is the reading practice, not the
model, that treats the output as script-neutral.

There is good reason to doubt the premise. A tokeniser cuts text into
subword pieces before a model sees it. Subword vocabularies are built
from corpora in which Latin script dominates, so they hold few whole
units for an Indic word and fall back on very short pieces. The same
content then costs more tokens, and each token carries less
meaning~\citep{DBLP:conf/nips/PetrovLB23,ahia-etal-2023-languages}. We
call this splitting \emph{fragmentation}. What fragmentation costs a
model is by now documented independently of evaluation. \citet{ghosh2026languagemodelsequallyrobust} test $27$
languages on six tasks. They find that robustness to alternative
tokenisations holds comfortably for English but fails to generalise.
Llama-3.1-8B loses $23.7\%$ of relative performance on average, and the
languages that fragment most suffer most. A learned metric is itself a multilingual
model, and it reads the same fragmented text. It is also trained and
validated mostly on the languages that fragment least. Fragmentation
degrades what a model does with a sentence. We ask whether it also
degrades what a metric says about one, and what a practitioner should do
if it does.

\noindent\textbf{Testing the premise:}
We put Eq.~\ref{eq:si} to a direct test on
IndicMT~Eval~\citep{DBLP:conf/acl/BDNKKKD23}, which pairs
$1{,}400$ MQM-annotated segments with each of five English$\to$Indic
pairs: Gujarati (GUJ), Tamil (TAM), Malayalam (MAL), Marathi (MAR) and
Hindi (HIN). Romanising the target side with a deterministic
transliterator holds source text, reference content and MQM annotation
fixed while varying orthographic form alone, so any movement in COMET is
attributable to script. Two gaps make the test worth running. Script has
rarely been treated as a controlled causal variable, and no
inference-time diagnostic exists that quantifies the resulting distortion
without retraining. The magnitudes involved are not subtle:
Table~\ref{tab:motivating} shows an error-free Malayalam output losing
more than $38$ COMET points to a change of script alone, and two
independently error-free translations of one English sentence separated
by $62.6$ points. \S\ref{sec:effect} reports the full results.

\begin{table}[t]
  \centering\small
  \setlength{\tabcolsep}{5pt}
  \begin{tabular*}{\columnwidth}{@{\extracolsep{\fill}}lcc@{}}
    \toprule
    \multicolumn{3}{@{}l}{(A) One translation, two scripts} \\[1pt]
    \multicolumn{3}{@{}p{0.9\columnwidth}}{\footnotesize
      \textit{``It was the final match for the All Blacks, who had
      already won the trophy two weeks ago.''}} \\[3pt]
    \textbf{Condition}      & \textbf{Human} & \textbf{COMET} \\
    \midrule
    MAL native script        & $25/25$        & $94.4$         \\
    Same text, romanised     & $25/25$        & $56.2$         \\
    \midrule
    \multicolumn{3}{@{}l}{(B) One source, two error-free targets} \\[1pt]
    \multicolumn{3}{@{}p{0.9\columnwidth}}{\footnotesize
      \textit{``Hydrogen ions are protons that had their electrons
      stripped off them.''}} \\[3pt]
    \textbf{Language / Script}     & \textbf{Human} & \textbf{COMET} \\
    \midrule
    MAL (Malayalam script)          & $25/25$        & $95.6$         \\
    HIN (Devanagari)                & $25/25$        & $33.0$         \\
    \bottomrule
  \end{tabular*}
  \caption{Two segments from IndicMT~Eval. Human columns are MQM
    annotations. Rewriting the script in (A) changes surface form alone,
    so the rating transfers unchanged; in (B) a single English source is
    rendered into two languages and both outputs are independently judged
    error-free, leaving script as the only source of variation.}
  \label{tab:motivating}
\end{table}

\noindent\textbf{What a measurement leaves open:}
Measuring the script-induced component of a score establishes that the
component exists and how large it is. It does not say what number to
publish in place of the affected one. Both obvious responses fail.
Discarding the score forfeits the evaluation. Publishing it unchanged
carries the confound into a system ranking. A third option is to romanise
every target, so that all languages are read in one script. That costs up
to $5.57\times$ more encoder compute, and it lowers agreement with
annotators instead of restoring it (\S\ref{sec:diag}, \S\ref{sec:rec}).
Earlier work does rescale metric outputs, but along the language axis
rather than the script axis, so it does not close the gap either
(\S\ref{sec:related}). Supplying the missing action is the purpose of
this paper.

\noindent\textbf{This paper:}
We begin from an observation that reframes the problem: what is loosely
called script bias is two failures wearing one name. Each (language,
script) pair forms a \emph{cell} (\S\ref{sec:norm}), and cells occupy
incompatible numerical ranges, so a Malayalam score and a Hindi score
cannot meaningfully be set side by side. We call this a failure of
\emph{calibration}. Separately, the metric grows
worse \emph{within} any single cell at telling better translations from
worse ones. Marathi is the clearest case. Its mean score barely moves
under romanisation, so the language looks well calibrated, yet its
ability to rank translations by quality, which we call
\emph{sensitivity}, falls as far as in any other language
(\S\ref{sec:marathi}). Each fault admits a different remedy. We prove
that the second is beyond the reach of every order-preserving transform
of the score, and we remove the first exactly using a construction
borrowed from high-throughput biology, with no human labels and no
retraining.

The paper is organised around three research questions. \textbf{RQ1:} how much
of a COMET score is attributable to the script of the target rather than
to the quality of the translation, and can that component be estimated at
inference time without human ratings (\S\ref{sec:effect})?
\textbf{RQ2:} scores from different scripts occupy different numerical
ranges. Can those ranges be made comparable by post-processing the scores
alone, without retraining the metric (\S\ref{sec:qn})? This matters
because
comparison across cells is itself a ranking task. Any multilingual
leaderboard, release threshold or average over language pairs pools
cells, and pooled agreement with annotators sits far below within-language
agreement for that reason (\S\ref{sec:impossibility}). \textbf{RQ3:} can
the lost within-cell sensitivity be restored
(\S\ref{sec:impossibility}, \S\ref{sec:sens})?

\noindent\textbf{Contributions:} We contribute (i)~the first controlled
causal test of Script Invariance for a learned MT metric, holding source,
reference content and human ratings fixed while varying script alone
(\S\ref{sec:effect}); (ii)~three inference-time diagnostics, built from
tokenisation and information parity, that quantify the script-induced
component of a score without human labels (\S\ref{sec:diag});
(iii)~a decomposition of script bias into calibration and sensitivity
components, shown to be empirically decoupled, with a severity-resolved
account of the Marathi counter-case (\S\ref{sec:marathi},
\S\ref{sec:decomp}); (iv)~a rank-invariance result bounding what any
post-hoc correction can achieve, demonstrated across four corrector
classes (\S\ref{sec:impossibility}); (v)~COMET-QN, a label-free
normalisation that removes the calibration component exactly and lifts
pooled cross-script agreement from $0.300$ to $0.399$ (\S\ref{sec:qn});
and (vi)~a parity-feature regressor recovering a further $17.1\%$ of lost
sensitivity with no exposure to the target language
(\S\ref{sec:sens}).

\section{Background and Related Work}
\label{sec:related}

\paragraph{Unequal tokenisation and parity measures:}
That subword vocabularies bill different languages differently for
equivalent content is by now well attested.
\citet{DBLP:conf/nips/PetrovLB23} document token-count disparities
reaching $15\times$ and introduce \textbf{Tokenization Parity} (TP), the
ratio of subword tokens per word in a target language against the English
figure; \citet{ahia-etal-2023-languages} follow the consequences through
to billing and latency on commercial APIs.
\citet{tsvetkov-kipnis-2024-information} recast the same disparity as
\textbf{Information Parity} (IP), the ratio between a language model's
compression efficiency on a target language and on English, and show IP
anticipates downstream behaviour better than token counts alone.
\citet{DBLP:conf/emnlp/KanjirangatSDR25} extend both measures to
dialectal settings. \citet{ghosh2026languagemodelsequallyrobust} report
the same coupling between fragmentation and model behaviour. It is
therefore not peculiar to any one architecture. Two vocabulary-independent quantities serve as controls
here. \textbf{Byte
Premium}~\citep{arnett-etal-2024-bit,arnett-bergen-2025-byte} is the
target-to-English ratio of bytes per word, and it eliminates UTF-8
overhead as an explanation.
\textbf{MATTR}~\citep{covington-mcfall-2010-mattr} counts distinct word
forms within a sliding $500$-token window. It reads no model vocabulary,
so it is immune to training-corpus artefacts
(Appendix~\ref{app:corp}). Together, TP and IP describe what we call the
\emph{representational burden} a tokeniser places on a language. TP
counts how many subword pieces the tokeniser spends on a given amount of
text, so it measures fragmentation. IP measures how much information each
of those pieces carries, so it measures density. Neither measure is
sufficient alone. A language can fragment little and still lose
information on every piece, or fragment heavily while each piece stays
meaningful. Only the two read together say what the tokeniser costs, so
the diagnostics of \S\ref{sec:diag} are built from their interaction.

This literature quantifies what a tokeniser costs a \emph{language}, and
\citet{ghosh2026languagemodelsequallyrobust} extend the accounting to
what it costs a \emph{model} at inference. Neither asks whether the cost
propagates into the \emph{evaluation scores} by which such models are
ranked. That is the question RQ1 puts.

\paragraph{Documented COMET failure modes:}
Work on where COMET breaks down has considered language mismatch and
translationese~\citep{DBLP:conf/wmt/ZouharBBN24} and perturbations
applied within a single script~\citep{karpinska-etal-2022-demetr}, and
has advocated retraining the encoder for under-served
pairs~\citep{DBLP:conf/naacl/WangAAMRBCHBBMO24,falcao-etal-2024-comet}.
None manipulates orthographic form as a controlled condition. The
IndicMT~Eval authors observed cross-language COMET variation and
attributed it to language family rather than testing
script~\citep{DBLP:conf/acl/BDNKKKD23}. Retraining, meanwhile, speaks to
neither the diagnosis nor the reporting difficulty confronting a
deployment already in service.

\paragraph{The COMET family shares one tokeniser:}
The checkpoint we audit (\S\ref{sec:setup}) inherits the
SentencePiece~\citep{kudo-richardson-2018-sentencepiece} tokeniser and
$250$K vocabulary of
XLM-R~\citep{DBLP:conf/acl/ConneauKGCWGGOZ20}. So does the rest of the
family. xCOMET-XL and xCOMET-XXL are trained over the XL and XXL variants
of XLM-R~\citep{guerreiro-etal-2024-xcomet,goyal-etal-2021-larger}, which
retain that vocabulary, and CometKiwi rests on
InfoXLM~\citep{rei-etal-2022-cometkiwi}, itself initialised from
XLM-R~\citep{chi-etal-2021-infoxlm}. TP is determined by the tokeniser
and not by the regression head, so the fragmentation burden we measure is
identical across these variants. The exposure travels with the vocabulary
rather than with the text, which is what makes a metric-side correction
the right level of intervention (\S\ref{sec:norm}).

\paragraph{Distributional correction and batch effects:}
Rescaling metric outputs is not itself novel, but it has been practised
along the language axis rather than the script axis: the COMET family
$z$-normalises annotations during
training~\citep{DBLP:conf/emnlp/ReiSFL20}, and low-resource variants
standardise direct assessments within
evaluator~\citep{DBLP:conf/naacl/WangAAMRBCHBBMO24}. Both are affine
maps, and \S\ref{sec:impossibility} proves that any affine map, indeed any
order-preserving transform, leaves a metric's rank agreement with human
judgement exactly unchanged, so it cannot repair the sensitivity half of
the present problem. Quantile normalisation is a more
forceful instrument, devised in high-throughput biology to remove
technical batch effects from microarray readings, and later refined for
cross-platform
comparison~\citep{bolstad-etal-2003-comparison,franks-etal-2018-feature}.
There, one biological sample measured on two machines yields two
readings that differ because of the equipment, and forcing both batches
onto a shared distribution removes the machine-specific component. The
correspondence we exploit is structural: a (language, script) cell stands
to a COMET score as a processing batch stands to a gene-expression
reading, with the writing system in the role of the machine, an artefact
of measurement rather than a property of the translation. So far as we
are aware, quantile
normalisation has not previously been enlisted to make learned-metric
scores comparable across writing systems.

\section{Experimental Setup}
\label{sec:setup}
This section fixes the corpus, the intervention, the checkpoint and the
sample bases.

\paragraph{Corpora:}
IndicMT~Eval~\citep{DBLP:conf/acl/BDNKKKD23} contributes $1{,}400$
MQM-annotated segments for each of the five English$\to$Indic pairs
introduced in \S\ref{sec:intro}. Each segment
carries fine-grained error types and one of five severity grades, running
from Very Low to Very High, alongside a \texttt{Default} grade for output
found free of error. Because one set of English sources is rendered into all five
targets, cross-language comparison rests on identical content. Two WMT24
pairs~\citep{kocmi-etal-2024-findings} serve as Latin-script controls:
ENG-SPA as a morphologically simple anchor and ENG-DEU as a
morphologically complex counterweight.\footnote{Obtained from
\url{https://github.com/google-research/mt-metrics-eval}}

\paragraph{The manipulation:}
Target-side text is transliterated by
IndicXlit~\citep{madhani-etal-2023-aksharantar}, a deterministic neural
system that leaves phonology, morphology and meaning intact while
altering orthographic form. Sources and MQM annotations are held constant,
so native-script judgements \emph{carry over} unchanged to the romanised
condition and any movement in COMET is ascribable to script alone.
Romanised Indic text is in any case an authentic
genre~\citep{roark-etal-2020-dakshina,chavan-etal-2023-boli}, so the
experiment doubles as an audit of metric behaviour on text deployments
genuinely meet.

\noindent\textbf{Transliteration quality:}
On Dakshina, a benchmark that pairs native-script words with human
romanisations~\citep{roark-etal-2020-dakshina}, IndicXlit reports a mean
top-1 accuracy of $60.5\%$, or $72.1\%$ with
re-ranking~\citep{madhani-etal-2023-aksharantar}. Both figures are for
the Roman-to-native direction. Exact match understates faithfulness here,
because romanisation is not standardised and several spellings of a word
are valid at once. The same error analysis attributes $85\%$ of errors to
vowel and consonant choice rather than to content. Residual noise is
therefore orthographic, and \S\ref{sec:restat} shows that noise of this
kind cannot account for the effect we measure.

\paragraph{Checkpoint and scope:}
Scores come from the reference-based checkpoint
\texttt{Unbabel/wmt22-comet-da}~\citep{rei-etal-2022-comet22} run under
\texttt{unbabel-comet} v2.2.6. TP uses XLM-R's SentencePiece tokeniser,
the one that checkpoint employs, and IP an external autoregressive model,
BLOOM-560m~\citep{lescao-etal-2023-bloom}.
Our empirical claims are bounded accordingly: to XLM-R-based COMET, and
to the five Indic abugidas studied, that is, scripts whose consonant
letters bear an inherent vowel modified by diacritics. The diagnostics
themselves read whichever tokeniser the audited metric carries, so the
apparatus transfers to xCOMET and CometKiwi unaltered. Other vocabularies
are taken up in the Limitations.

\paragraph{Three sample bases:}
The corpus holds $5\times1{,}400=7{,}000$ segments, and three nested
subsets are used, identified wherever a figure is quoted.
\textbf{(i)~All $7{,}000$ segments} underpin every analysis using
metric-side quantities alone: the ANOVA, every COMET, TP and IP mean, the
parity diagnostics of \S\ref{sec:diag}, and the Hindi--Gujarati
comparison of \S\ref{sec:obs}. \textbf{(ii)~A $6{,}995$-segment
working set} underpins every analysis requiring a human score, five rows
bearing non-numeric MQM entries being discarded. At most $0.2\%$ of any
language is touched and no conclusion turns on the choice, as
Appendix~\ref{app:stats} shows. \textbf{(iii)~Severity-resolved subsets}
exclude \texttt{Default}, which bears no severity rank, leaving
$1{,}258$ segments for MAR (\S\ref{sec:marathi}). Counts appear in
Appendix~\ref{app:stats}.

\paragraph{Conventions:}
Tables and figures order languages by native-script COMET descending,
which for this corpus gives GUJ, TAM, MAL, MAR, HIN, running from $85.7$
down to $71.8$. A dagger ($\sig$) marks a coefficient individually
significant against zero at $p<0.05$, two-sided and at segment level, and
$\nsig$ one that is not; each coefficient is tested alone, and
differences between two dependent coefficients are tested separately
(\S\ref{sec:restat}). Subscripts $_\mathrm{nat}$ and $_\mathrm{rom}$
denote the native-script and romanised conditions. With
the apparatus fixed, we turn to what changing the script does to the
score.

\section{The Effect of Romanisation}
\label{sec:effect}
Romanisation changes the script and nothing else, so whatever moves in
the score is the script moving it. This section records what moves, in
three observations. It then turns those observations into diagnostics a
practitioner can compute, and asks whether the decline they describe is
large enough and real enough to warrant a correction. Supporting tables
sit in Appendices~\ref{app:stats} to \ref{app:corp}.

\begin{table*}[t]
\centering\small
\setlength{\tabcolsep}{5pt}
\begin{tabular*}{\textwidth}{@{\extracolsep{\fill}}l rrrc rrr rrr@{}}
\toprule
\textbf{Language} & \multicolumn{4}{c}{\textbf{COMET}}
  & \multicolumn{3}{c}{\textbf{TP}} & \multicolumn{3}{c}{\textbf{IP}} \\
\cmidrule(r){2-5} \cmidrule(r){6-8} \cmidrule(r){9-11}
& nat & rom & $\Delta$ & $\rho_\mathrm{nat}{\to}\rho_\mathrm{rom}$
& nat & rom & $\Delta$ & nat & rom & $\Delta$ \\
\midrule
GUJ$^{a}$ & $85.70$ & $70.61$ & $-15.09$ & $0.590\sig\!\to\!0.341\sig$
  & $1.388$ & $1.599$ & $+15.2\%$ & $0.438$ & $0.289$ & $-34.0\%$ \\
TAM       & $84.88$ & $77.72$ & $-7.15$  & $0.629\sig\!\to\!0.240\sig$
  & $1.319$ & $2.347$ & $+77.9\%$ & $0.545$ & $0.174$ & $-68.0\%$ \\
MAL       & $84.08$ & $70.07$ & $-14.01$ & $0.651\sig\!\to\!0.460\sig$
  & $1.248$ & $1.847$ & $+48.0\%$ & $0.549$ & $0.197$ & $-64.1\%$ \\
MAR$^{b}$ & $71.92$ & $72.94$ & $+1.02$  & $0.487\sig\!\to\!0.278\sig$
  & $1.165$ & $1.640$ & $+40.7\%$ & $0.490$ & $0.280$ & $-42.9\%$ \\
HIN       & $71.75$ & $68.32$ & $-3.43$  & $0.422\sig\!\to\!0.230\sig$
  & $1.210$ & $1.702$ & $+40.7\%$ & $0.636$ & $0.265$ & $-58.3\%$ \\
\bottomrule
\end{tabular*}
\caption{Mean COMET, segment-level Spearman agreement with MQM, and the
  TP--IP inversion, in the native (nat) and romanised (rom) conditions.
  COMET, TP and IP means use all $1{,}400$ segments per language;
  $\rho$ uses the working set. $^{a}$GUJ shows the largest COMET fall
  despite the smallest TP inflation, because its native IP is the lowest
  of the five, which gives it the greatest representational burden
  (\S\ref{sec:diag}).
  $^{b}$MAR is the sole language whose mean COMET rises while every other
  quality signal worsens (\S\ref{sec:marathi}).}
\label{tab:results}
\end{table*}

\subsection{Three observations}
\label{sec:obs}

\noindent\textbf{Observation 1: script explains $22.9\%$ of COMET
variance.} A one-way ANOVA grouping targets by script family
(Devanagari: HIN, MAR; non-Devanagari: GUJ, MAL, TAM) gives
$\eta^2=22.9\%$ on native-script COMET, and romanisation reduces it to
$1.5\%$ (Appendix~\ref{app:stats}). The effect is concrete: Hindi and
Gujarati are two Indo-Aryan languages separated chiefly by writing
system, and the $13.95$-point COMET gap between them shrinks to $2.29$
under romanisation. Script is therefore the dominant factor behind
cross-language COMET differences. The displacement is specific to each
language and does not even share a sign: GUJ falls by $15.09$ points,
MAL by $14.01$, TAM by $7.15$ and HIN by $3.43$, while MAR \emph{rises}
by $1.02$ (Table~\ref{tab:results}). A spread of that width, with one
language moving against the other four, already rules out any correction
written as a single constant or slope (\S\ref{sec:decomp}).

\noindent\textbf{Observation 2: the TP--IP inversion.} Romanisation
inflates TP and collapses IP in every Indic language, TAM at the extreme
with TP $+77.9\%$ and IP $-68.0\%$ (Table~\ref{tab:results}). At the
language level ($n=5$, descriptive) the TP--IP anti-correlation
strengthens from $r=-0.481\nsig$ natively to $r=-0.896\sig$ under
romanisation. The explosion is driven by Tamil digraphs (\textit{zh},
\textit{dh}, \textit{gh}) and by Indic dependent vowels that the
tokeniser strands as near-zero-information
pieces~\citep{brahma-etal-2025-morphtok}, a mechanism examined in
Appendix~\ref{app:dv}. The encoder is handed more tokens each carrying
less, the worst combination for a signal that has to discriminate.

\noindent\textbf{Observation 3: COMET is uniquely fragile.} In native
script COMET has the highest $|\rho|$ against MQM in every language,
ranging from $0.422$ to $0.651$. Under romanisation it falls furthest in
every language ($|\Delta\rho|=0.19$ to $0.39$), exceeding
BLEURT~\citep{DBLP:conf/acl/SellamDP20} at $0.09$ to $0.19$,
BERTScore~\citep{DBLP:conf/iclr/ZhangKWWA20} at $0.04$ to $0.06$, and the
surface metrics BLEU, ChrF~\citep{DBLP:conf/wmt/Popovic15} and
TER~\citep{snover-etal-2006-ter}, all below $0.07$;
Table~\ref{tab:fullcorr} gives all six. Agreement falls in all five
languages, MAR included, so the displacement is not a harmless
relabelling of a scale that still orders translations correctly. Every
metric reads the same text against the same ratings, and only the
encoder-dependent ones move, which places the failure inside the encoder
rather than at the character surface. These observations document the
failure but give a practitioner nothing to compute, which is what the
next subsection supplies.

\subsection{The parity diagnostics}
\label{sec:diag}
We now introduce three diagnostics that turn the observations into
numbers computable before any score is trusted. TP and IP each capture one
dimension of the representational burden of \S\ref{sec:related}, and the
Gujarati case shows why the two must be combined. Formal definitions, axiomatic motivation and threshold
calibration are in Appendices~\ref{app:langprof} and \ref{app:axioms}.

\noindent\textbf{Script Bias Index:}
$\mathrm{SBI}=\mathbb{E}_i[\mathrm{TP}_i/\mathrm{IP}_i]$ counts subword
pieces consumed per unit of information retrieved. Both inputs are cheap.
Token counts come from the tokeniser. The information term comes from
BLOOM-560m~\citep{lescao-etal-2023-bloom}, a small multilingual language
model used here only to measure how predictable a segment is. SBI
therefore never reads COMET output or human ratings, and can be computed
before any evaluation is trusted. We take $\mathrm{SBI}\ge3.0$ as a
working threshold. Above it we do not read a raw score as a quality
signal without the corrections of \S\ref{sec:norm}. The native-script
ranking of SBI over the five languages reproduces the COMET ranking
exactly ($\rho=1.00$).

SBI also settles a case that neither parity measure settles alone, which
we call the Gujarati anomaly. Gujarati fragments least of the five under
romanisation, its TP rising by only $+15.2\%$, so fragmentation on its
own would call it the safest language here. Its native information parity
is nonetheless the lowest at $0.438$, meaning each of its tokens carries
the least. Combining the two gives $\mathrm{SBI}_\mathrm{nat}=3.45$, the
highest of the five, and that matches how COMET behaves: Gujarati loses
more points under romanisation than any other language
(Table~\ref{tab:results}). Romanisation raises SBI sharply in every
language, from
$\mathrm{SBI}_\mathrm{nat}\in[2.03,3.45]$ to
$\mathrm{SBI}_\mathrm{rom}\in[5.96,15.90]$, and the flag fires on at
least $95.2\%$ of romanised segments in all five languages against
$6.6\%$ to $62.1\%$ natively.

\noindent\textbf{IP Parity Index:}
$\mathrm{IPI}=|\mathrm{IP}-1.0|$ is distance from English-equivalent
parity, and it sorts a setting into three zones. In \textbf{Parity},
below $0.35$, scores may be read directly. In \textbf{Burden}, between
$0.35$ and $0.70$, COMET still tracks adequacy but its absolute level is
suppressed. In \textbf{Paradox}, above $0.70$, agreement with MQM typically falls
below $\rho=0.40$; romanised MAL is the sole exception
(Appendix~\ref{app:axioms}). All five Indic languages sit in Burden
natively and cross into Paradox once romanised.

\noindent\textbf{Computational Tax:}
Romanisation charges twice at once, in tokens per word
($\mathrm{LP}=\overline{\mathrm{TP}}_\mathrm{rom}/\overline{\mathrm{TP}}_\mathrm{nat}$)
and in information per token
($\mathrm{EP}=\overline{\mathrm{IP}}_\mathrm{nat}/\overline{\mathrm{IP}}_\mathrm{rom}$),
and since neither charge can be traded for the other their product
$\mathrm{Tax}=\mathrm{LP}\times\mathrm{EP}$ gives the surplus encoder
compute a romanised segment demands, from $1.75\times$ for GUJ to
$5.57\times$ for TAM. EP accounts for $62$ to $75\%$ of
$\ln(\mathrm{Tax})$ (Appendix~\ref{app:tax}), so the dominant cost is
information lost per token, which no remedy that merely counts tokens can
reach.

Between them the three answer RQ1. They estimate the script-induced
component of a score from token counts and likelihoods alone, and every
romanised setting studied here exceeds both working thresholds,
$\mathrm{SBI}\ge3.0$ and $\mathrm{IPI}>0.70$. What they do not supply is
a number to publish in place of the affected one. Before building one, the
next subsection asks whether the decline they flag is large enough and
real enough to warrant a correction at all.

\subsection{Effect size and the adequacy objection}
\label{sec:restat}
Before building a correction we settle two questions. Is the decline real
at a defensible effect size, and is it instead the metric responding
correctly to a genuine loss of adequacy?

\noindent\textbf{Measuring the effect on the right scale:}
A correlation coefficient does not live on a ratio
scale~\citep{stevens-1946-theory}, so a claim of the form ``the
correlation dropped by $X$ percent'' has no defined referent. We
therefore report absolute differences, and shared variance where a
relative statement is wanted. COMET's agreement with MQM declines by
$|\Delta\rho|$ between $0.19$ and $0.39$; expressed as shared variance,
$\rho^2$ falls from $0.42$ to $0.21$ for MAL and from $0.40$ to $0.06$
for TAM. The two coefficients are computed over the same segments against the same
ratings, so they are dependent and overlapping, and the difference is
tested with the \citet{meng-etal-1992-comparing} procedure. Every
language shows a significant decline, with $z$ running from $8.7$ to
$16.0$ and all $p<10^{-17}$. Appendix~\ref{app:stats} gives the
per-language values, two confirmatory tests, and the relation to the
Williams test used for metric comparison at WMT.

\noindent\textbf{Is adequacy genuinely lost?}
Because MQM ratings are carried over rather than re-collected, one might
object that romanisation truly degrades adequacy and that a falling
correlation is the metric behaving correctly. Two features of the data
resist that reading. First, all six metrics are scored against the same
ratings, yet the size of the decline tracks how far each metric depends
on a subword encoder (Appendix~\ref{app:stats}), whereas invalid ratings
would drag every metric down alike. Second, \S\ref{sec:marathi} contains
outputs whose COMET scores \emph{rise} under romanisation while their
human adequacy is unchanged and, in one instance, zero; a translation
judged entirely inadequate cannot be made more adequate by a change of
script.
The same two features answer the related worry that transliteration
itself introduces errors. Noise of that kind would act like annotation
noise, and would therefore depress every metric alike rather than
tracking encoder dependence. Direct human rating of romanised output
would settle both questions outright and remains for future work. A correction is therefore
warranted, and the next section establishes what kind it must be.

\section{The Marathi Severity Inversion}
\label{sec:marathi}
Marathi behaves unlike the other four languages, and the reason
constrains what any correction can look like. Romanisation lowers mean
COMET in the other four, which at least moves with their falling
agreement with human judgement. Marathi does the opposite: its mean COMET
\emph{rises} by $1.02$ points, even though every underlying quality
signal worsens in step with the rest, IP falling by $42.9\%$ and
agreement with MQM dropping from $\rho=0.487$ to $0.278$. Observation~1 recorded the rise. This section accounts for it,
and the account carries beyond Marathi, because it eliminates the most
obvious class of remedy before any is built.

The immediate cause is orthographic. Marathi marks spoken and literary
variation through diacritics including visarga, a word-final breathing
mark, and chandrabindu, a nasalisation
mark~\citep{ravishankar-tyers-2017-finite}. Romanisation collapses these
distinctions onto one Latin surface form, rendering every hypothesis more
reference-like to a similarity-based metric regardless of its quality.
Resolving by MQM severity exposes the consequence
(Table~\ref{tab:marathi-inversion}). The gain accrues entirely to badly
translated segments while the best output loses ground, and bucket means
rise monotonically with severity. At segment level, severity and
$\Delta$COMET correlate at $\rho=0.183\sig$ ($p=5.9\times10^{-11}$,
$N=1{,}258$). The aggregate rise is significant across the full $1{,}400$ (Wilcoxon
$W=453{,}856$, $p=0.016$), yet the median change is $-0.21$: a minority
of very poor segments pulls the mean upward. One output rated $0/25$
gains $26.1$ COMET points, from $36.6$ to $62.7$.

Two things follow. Evidentially, and as used in \S\ref{sec:restat}, score
movements that reward inadequacy cannot be evidence of recovered
adequacy. Prescriptively, Marathi's bias is neither a fixed offset nor a
fixed multiple: it varies with quality within the language, and its
aggregate sign opposes that of the other four. No single constant can
therefore serve all five. That rules out the per-language offsets and
rescalings a practitioner would reach for first, and leaves only a
treatment acting on whole score distributions rather than on their
centres. Constructing one is the business of \S\ref{sec:norm}, and
\S\ref{sec:impossibility} makes the exclusion exact.

\begin{table}[t]
  \centering
  \small
  \setlength{\tabcolsep}{0pt}
  \begin{tabular*}{\columnwidth}{@{\extracolsep{\fill}}lrr@{}}
    \toprule
    \textbf{Severity} & \textbf{\(N\)} & \textbf{Mean \(\Delta\)COMET} \\
    \midrule
    Very Low  & 69  & \(-3.72\) \\
    Low       & 86  & \(-1.90\) \\
    Medium    & 184 & \(+0.73\) \\
    High      & 258 & \(+1.08\) \\
    Very High & 661 & \(+2.63\) \\
    \midrule
    \multicolumn{3}{@{}l}{\footnotesize%
      Segment-level severity--\(\Delta\)COMET:
      \(\rho = 0.183\sig\), \(N = 1{,}258\)} \\
    \bottomrule
  \end{tabular*}
\caption{Mean $\Delta$COMET (romanised minus native) for MAR by MQM
  severity. The five buckets total $1{,}258$; the remaining $142$ segments
  are \texttt{Default} and bear no severity rank. $\Delta$COMET needs no
  human score, so the full $1{,}400$-segment base applies.}
\label{tab:marathi-inversion}
\end{table}

\section{From Diagnosis to Treatment}
\label{sec:norm}
A practitioner arrives at this point holding a flagged score, that is,
one computed in a setting where $\mathrm{SBI}\ge3.0$ or
$\mathrm{IPI}>0.70$ (\S\ref{sec:diag}). This section supplies the missing
action in three moves: separate the bias into parts that demand
different remedies (\S\ref{sec:decomp}), establish what no post-hoc
method can achieve (\S\ref{sec:impossibility}), and correct what remains
(\S\ref{sec:qn} and \S\ref{sec:sens}). The correction is applied to the
scores rather than to the text because the fragmentation driving the
confound belongs to the vocabulary the metric carries
(\S\ref{sec:related}). Throughout, a \emph{cell} denotes one (language,
script) pair, such as GUJ native or GUJ romanised, and the ten settings
form a $5\times2$ grid of cells. A cell plays the role of a batch in the
quantile-normalisation analogy of \S\ref{sec:related}. Every quantity
here is validated against human ratings, so all values use the
$6{,}995$-segment working set (\S\ref{sec:setup}).

\subsection{Calibration and sensitivity are different problems}
\label{sec:decomp}
Marathi showed that no constant serves every language. The reason is that
romanisation does two things at once, and conflating them has obscured
which of the two is fixable.

\textbf{(P1) Miscalibration.} Cells occupy different numerical ranges.
The native-minus-romanised COMET gap varies across languages and even
reverses sign, standing at $+15.09$ points for GUJ but $-1.02$ for MAR
(\S\ref{sec:obs}), so no shared constant, slope or offset can remove it.

\textbf{(P2) Sensitivity loss.} Within a cell, the metric's capacity to
order translations by quality deteriorates. Segment-level agreement with
MQM falls in every language, MAR included, whose P1 gap is nearly zero.

P1 is a question of location and scale, P2 one of representational
fidelity, and MAR shows that the two vary independently: a cell can be
almost perfectly calibrated and still have lost much of its discriminative
power. The split is worth making because the two admit different
treatments, and, as the next subsection shows, one admits none at all.

\subsection{Rescaling cannot restore sensitivity}
\label{sec:impossibility}
This subsection fixes the limit on RQ3. P2 is beyond \emph{any} rescaling
of the score, so a correction for it must look outside the score entirely
(\S\ref{sec:sens}). Let $s=(s_1,\dots,s_n)$ be the COMET scores in a cell
and $h$ the paired human ratings, and recall that Spearman's $\rho_S$
depends on the scores only through their ranks.

\noindent\textbf{Proposition (Rank invariance):} For any strictly
increasing $f$, order is preserved, so
$\mathrm{rank}(f(s_i))=\mathrm{rank}(s_i)$ for every $i$, and therefore
\begin{equation}
\rho_S\bigl(f(s),\,h\bigr)=\rho_S\bigl(s,\,h\bigr).
\label{eq:rankinv}
\end{equation}

The two sides of Eq.~\ref{eq:rankinv} are equal for every dataset and
every strictly increasing $f$, not merely on average. The reason is that
such an $f$ can stretch or shift the scores but cannot reorder them, and
Spearman agreement depends on the scores only through their order.
Rescaling therefore leaves agreement with human judgement exactly where
it was. The scope of that argument is what makes it useful: mean-shifting, affine rescaling, quantile mapping
and isotonic regression are all strictly increasing functions of the
scalar score, and
together they exhaust the class of post-hoc corrections a practitioner
would plausibly reach for. Whatever discriminative power romanisation
destroyed at scoring time survives every one of them untouched.

Table~\ref{tab:impossibility} confirms this on a 50/50 within-language
split. Four distributional correctors reduce the mean gap from $7.88$ COMET
points on the held-out half to roughly $0.6$, and they lift pooled
cross-language agreement. The whole-working-set figure is $8.14$
(Figure~\ref{fig:qn}). Yet the language-averaged Spearman stays pinned at
$0.310$ against a native ceiling of $0.564$, moving in the fourth decimal
at most. The table also answers the objection that only rankings
matter and a per-script scale is therefore harmless. Pooled agreement
across cells is itself a ranking task, and on the raw romanised half it
stands at $0.160$ against $0.310$ within language. Every monotone
corrector lifts the pooled figure toward $0.30$ without moving the
within-language figure, which is the calibration fault seen as a ranking
failure.

The result is negative in form and constructive in consequence. It sends
effort where it can pay: P1 is fully remediable once the scores exist
(\S\ref{sec:qn}), P2 only by using information beyond the scalar
(\S\ref{sec:sens}), and the rest is the Computational Tax of
\S\ref{sec:diag}.

\begin{table}[t]
\centering\small
\setlength{\tabcolsep}{4pt}
\resizebox{\columnwidth}{!}{%
\begin{tabular}{@{}lccc@{}}
\toprule
\textbf{Corrector (transform class)} & $|\Delta|$
  & $\rho_S^{\mathrm{lang}}$ & $\rho_S^{\mathrm{pool}}$ \\
\midrule
Raw romanised (identity)            & $7.88$ & $0.310$ & $0.160$ \\
Per-lang mean shift (location)      & $0.61$ & $0.310$ & $0.280$ \\
Per-lang affine / $z$ (loc.+scale)  & $0.63$ & $0.310$ & $0.301$ \\
Per-lang quantile map (monotone)    & $0.61$ & $0.310$ & $0.298$ \\
Per-lang isotonic (monotone)        & $0.68$ & $0.309$ & $0.282$ \\
Native score (ceiling)              & $0.00$ & $0.564$ & $0.488$ \\
\bottomrule
\end{tabular}}
\caption{Rank invariance quantified on a 50/50 within-language
  split (seed 42), so all values are computed on the held-out half:
  $|\Delta|$ is the mean absolute per-language gap after correction (the
  whole-working-set raw gap is $8.14$, Figure~\ref{fig:qn}),
  $\rho_S^{\mathrm{lang}}$ the language-averaged and
  $\rho_S^{\mathrm{pool}}$ the pooled agreement with MQM over the
  romanised cells of that half. Every monotone corrector closes the gap
  and improves pooled comparability while the language-averaged column
  remains fixed at $0.310$, exactly as Eq.~\ref{eq:rankinv} requires.}
\label{tab:impossibility}
\end{table}

\subsection{COMET-QN: quantile renormalisation}
\label{sec:qn}
This subsection answers RQ2. The calibration fault P1 can be removed
exactly, and the idea behind the removal is simple. We force every cell onto one shared score distribution, so
that a given rank means the same COMET value in any language or script.
We take the reference $R$ to be the pooled native-script distribution,
and for a score $s_i$ in a cell of size $n$ we set
\begin{equation}
s'_i = Q_R\!\left(\frac{\mathrm{rank}(s_i)}{n+1}\right),
\label{eq:qn}
\end{equation}
with $Q_R$ the empirical quantile function of $R$. This is quantile
normalisation as used to remove batch effects from microarray
readings~\citep{bolstad-etal-2003-comparison}, applied to cells instead
of batches. Three steps: keep only each score's rank, turn it into a
position in $[0,1]$, and read off the value at that position of $R$. Appendix~\ref{app:qnproc} works five
romanised Gujarati segments through the map.

\begin{figure*}[t]    
  \centering
  \includegraphics[width=\textwidth]{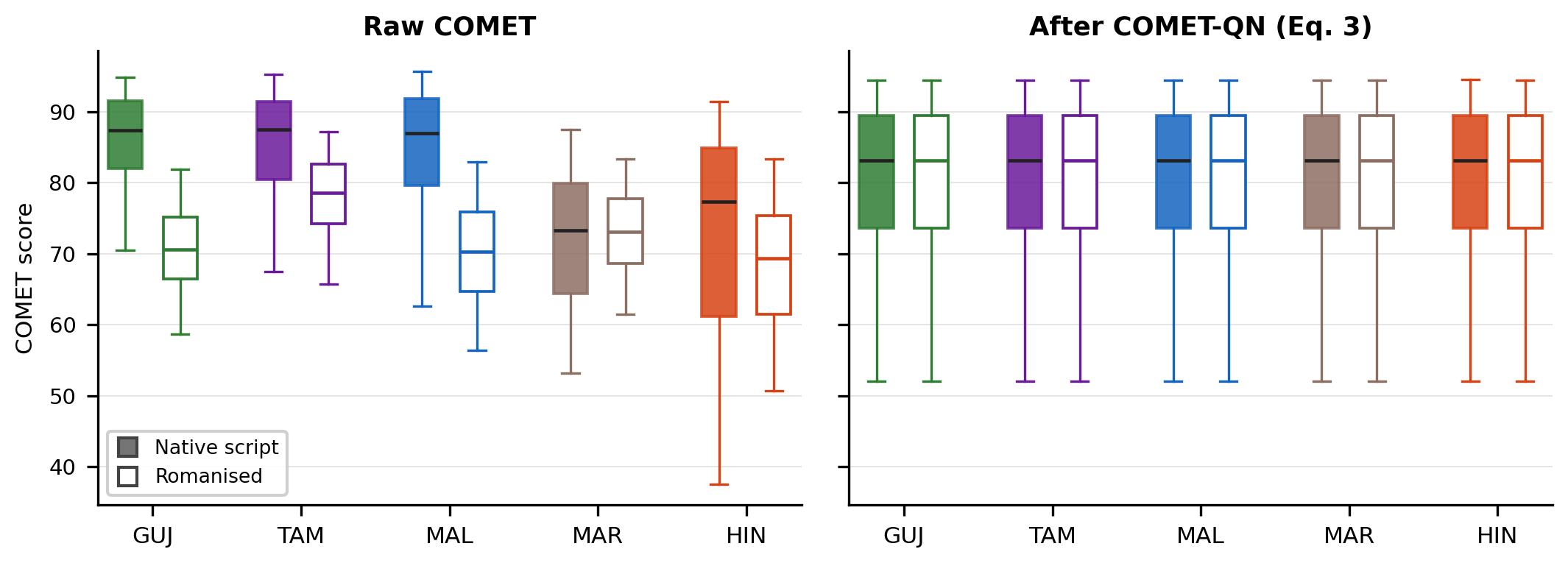}
  \caption{COMET-QN across the ten (language, script) cells (boxes:
    median, IQR, 5th--95th percentiles; $6{,}995$-segment working set).
    \textbf{Left, raw COMET} (mean $|$per-lang.\ gap$|=8.14$ pts; pooled
    $\rho_S=0.300$): native cells, filled, sit consistently above their
    romanised counterparts, open, by a margin that varies across
    languages and reverses for MAR, which is problem P1
    (\S\ref{sec:decomp}). \textbf{Right, after COMET-QN}
    (Eq.~\ref{eq:qn}; mean $|$per-lang.\ gap$|=0.00$ pts; pooled
    $\rho_S=0.399$): all cells share the pooled native reference, so the
    gap is exactly zero by construction and cross-script comparison
    becomes meaningful. Pooled Pearson agreement rises from $0.324$ to
    $0.400$ over the same cells, while within-language agreement is
    unchanged to four decimals, GUJ standing at $0.5903$ before and
    after. Within-cell ordering is provably untouched
    (Eq.~\ref{eq:rankinv}).}
  \label{fig:qn}
\end{figure*}

Because Eq.~\ref{eq:qn} uses $s_i$ only through its rank, it cannot
disturb any within-cell ordering (Eq.~\ref{eq:rankinv}), yet every cell
now shares $R$, so the per-language gap, MAR's sign-reversed $-1.02$
included, is exactly zero by construction rather than by fitting.
Figure~\ref{fig:qn} confirms the effect. The gap collapses to zero.
Pooled agreement with MQM rises from $0.300$ to $0.399$ (Spearman) and
$0.324$ to $0.400$ (Pearson), while within-language agreement is
unchanged to four decimals. COMET-QN buys
comparability, not sensitivity, exactly as the decomposition predicts.
SBI also collapses toward zero afterwards, so the diagnostic of
\S\ref{sec:diag} doubles as a certificate that P1 is eliminated.

\noindent\textbf{Scope and reference:} COMET-QN assumes the cells share a
comparable underlying quality distribution. If one system pool is
genuinely weaker, the method erases that real difference along with the
artefactual one, so it belongs within a single dataset and domain. The
same restriction governs comparison across test sets. Two test sets with
different quality distributions are different batches, so COMET-QN should
be applied within each and its outputs not compared between them;
quantifying the distortion when that assumption is violated is future
work. Stable quantiles need a few hundred segments per cell, easily met
by the $1{,}400$ here. The pooled native-script reference is the condition
COMET was validated on and hence the natural anchor. Other references
preserve the exact zero-gap property, though we do not claim the
pooled-agreement gain is invariant to the choice. P1 is now settled, which leaves P2 and the question of how
much of it can be won back.

\subsection{Recovering sensitivity from parity features}
\label{sec:sens}
RQ3 concerns P2, the sensitivity fault. It lies beyond any transform of
the scalar (Eq.~\ref{eq:rankinv}), so recovering it needs information the
scalar does not carry. The parity diagnostics are the natural source,
since they record how severely the encoder was taxed on each segment. We
therefore compose two operators,
\begin{equation}
\tilde{s} = C_{\mathrm{sens}}\bigl(C_{\mathrm{cal}}(s, c);\,\phi\bigr),
\label{eq:compose}
\end{equation}
read inside out: COMET-QN ($C_{\mathrm{cal}}$) fixes the calibration
fault while leaving every rank untouched, then the regressor
($C_{\mathrm{sens}}$) uses the parity features $\phi$ to re-rank
segments. Calibration must come first because it is the only rank-safe
step, and $C_{\mathrm{sens}}$ is deliberately the only stage allowed to
change ranks. Here $\phi$ collects TP and IP in both conditions together
with their differences, and $C_{\mathrm{sens}}$ is a gradient-boosted
regressor mapping $(\mathrm{COMET}_\mathrm{rom}, \phi)$ to
$\mathrm{COMET}_\mathrm{nat}$. No human labels enter: because
romanisation preserves content, every (romanised, native) pair is free
supervision.

Evaluation follows the honest deployment regime, leave-one-language-out:
fit on four languages and apply to the fifth, unseen, as when a new
language arrives with no corrector data of its own. The deficit recovered
is
$(\rho_{\mathrm{cal}}-\rho_{\mathrm{raw}})/(\rho_{\mathrm{nat}}-\rho_{\mathrm{raw}})$,
a scale that stays comparable across languages with different native
ceilings. Averaged over the five folds the regressor restores $17.1\%$ of the lost
sensitivity with no exposure to the target language, improving four of
five (Table~\ref{tab:lolo}).

\noindent\textbf{Reading the failure:} MAR's negative transfer at
$-20.3\%$ is the same fact that defeats a universal formula: script bias
is not one function. The parity features act as a proxy for the direction
of the bias, dependable only where the pattern of TP and IP divergence
resembles that of the training languages. The failure can nonetheless be
anticipated without labels, though not by the mean diagnostics: MAR's
native SBI of $2.61$ and IPI of $0.510$ both sit inside the range of the
other four (Appendix~\ref{app:langprof}). What separates MAR is the sign
of its mean cell gap, native minus romanised, computable from scores
alone and the direction the regressor learns. A cell whose sign opposes
the training cells should therefore receive COMET-QN alone, until a
corrector has been fitted on languages sharing its sign. The residual five sixths of the deficit is
encoder-intrinsic, which is to say it is the Computational Tax of
\S\ref{sec:diag}. Post-hoc composition and encoder retraining therefore
address disjoint portions of the loss, and neither substitutes for the
other.

What this buys a practitioner is modest but definite. Scores from
different scripts become comparable, a sixth of the lost discrimination
comes back, and the distortion that remains can be stated rather than
hoped away. The rest cannot be reported around, only rebuilt around, and
the procedure below is built on that division.

\begin{table}[t]
\centering\small
\setlength{\tabcolsep}{5pt}
\begin{tabular*}{\columnwidth}{@{\extracolsep{\fill}}lcccr@{}}
\toprule
\textbf{Held-out} & $\rho_{\mathrm{raw}}$ & $\rho_{\mathrm{cal}}$
  & $\rho_{\mathrm{nat}}$ & \textbf{Recovered} \\
\midrule
GUJ & $0.341$ & $0.437$ & $0.590$ & $38.6\%$ \\
TAM & $0.240$ & $0.291$ & $0.629$ & $13.0\%$ \\
MAL & $0.460$ & $0.524$ & $0.651$ & $33.5\%$ \\
MAR & $0.278$ & $0.236$ & $0.487$ & $-20.3\%$ \\
HIN & $0.230$ & $0.270$ & $0.422$ & $20.8\%$ \\
\midrule
Mean & \multicolumn{3}{c}{} & $17.1\%$ \\
\bottomrule
\end{tabular*}
\caption{Parity-feature calibration under leave-one-language-out
  (segment-level Spearman against MQM, $6{,}995$-segment working set).
  Recovery is the fraction of the raw-to-native sensitivity gap closed.
  MAR transfers negatively: its bias runs counter to the four training
  languages, its P1 gap being sign-reversed (\S\ref{sec:decomp}), so a
  model fitted on them mispredicts its residual.}
\label{tab:lolo}
\end{table}

\section{What to Report}
\label{sec:rec}
The preceding sections converge on a procedure. For a non-Latin-script
target we suggest the following.

\begin{enumerate}[leftmargin=1.3em,itemsep=0pt,topsep=1pt,parsep=0pt]
\item State which encoder and tokeniser the metric carries. The COMET
  variants in common use carry XLM-R (\S\ref{sec:related}); others must
  be profiled against their own vocabulary first.
\item Compute IPI once for the (language, script) pair against that
  tokeniser. This needs no human ratings and no access to the metric's
  internals.
\item In the Parity zone, report raw COMET. The tokeniser treats the
  language much as it treats English, so no companion figures are
  needed.
\item In the Burden zone, report raw COMET with IPI beside it, so
  absolute levels are not over-read while rankings are still trusted.
\item In the Paradox zone, report COMET-QN (Eq.~\ref{eq:qn}) with raw
  COMET and the three diagnostics, and do not compare raw scores across
  scripts. Apply the parity-feature regressor only when the cell's
  mean-gap sign matches the languages it was fitted on
  (\S\ref{sec:sens}).
\item Do not romanise as a preprocessing convenience: it costs
  $1.75$--$5.57\times$ in encoder compute and degrades agreement with
  annotators rather than equalising it (\S\ref{sec:restat}).
\end{enumerate}

\noindent No step consumes human ratings.
Appendix~\ref{app:worked} runs the procedure through one setting,
romanised Gujarati, end to end.

\section{Conclusion}
\label{sec:conc}
Script bias in COMET is two faults. One is removed exactly and without
labels by quantile renormalisation; the other is beyond any rescaling and
comes back only in part. The recommendation is concrete: publish the
normalised score, the diagnostics, and the identity of the tokeniser they
were computed against, whenever COMET is reported on a non-Latin-script
target. That tokeniser is XLM-R across the COMET variants in common use
(\S\ref{sec:related}), so the procedure covers much of published Indic MT
evaluation. The released code emits all four numbers in one pass, so
adopting the protocol costs a practitioner one command.

Two consequences reach past this corpus. A metric can look healthy in
aggregate and still be unfit for the comparison it is put to, so the
conditions a score was measured under belong beside it. The fault is also
not COMET's alone: it lives in a vocabulary much of the multilingual
metric family shares. Three things follow. The apparatus should be run against metrics
carrying other vocabularies, which is the study we are preparing. The
diagnostics should be tested on writing systems that are not abugidas,
once a corpus with matched sources exists. And human ratings should be
collected directly on romanised output, which would settle the one
assumption this design cannot test from the inside. The deeper remedy
remains an encoder that spreads capacity evenly across scripts.

\section*{Limitations}
\label{sec:lim}
Our metric-level claims cover one metric,
COMET~\texttt{wmt22-comet-da}, and one tokeniser, the XLM-R
SentencePiece vocabulary that checkpoint carries. The xCOMET and
CometKiwi variants inherit the same tokeniser and vocabulary
(\S\ref{sec:related}), so the fragmentation burden we measure carries to
them unchanged; whether the score-level distortion does is a different
question, and one we have not answered. Nor have we measured how
tokenisers of other designs, byte-level, English-centric or WordPiece,
respond to the same intervention, or how metrics built on them would
score the resulting text. Both questions belong to a single follow-up
study, which we are preparing: it takes the neural metrics that do
\emph{not} share XLM-R's vocabulary, MetricX~\citep{juraska-etal-2024-metricx} and the mT5-based
family~\citep{xue-etal-2021-mt5} most obviously, computes the parity diagnostics against each metric's own
tokeniser, and repeats the romanisation intervention across that set.
That design would establish whether script sensitivity is a property of
one widely-used vocabulary or a recurring hazard of multilingual metric
construction, and it would test whether COMET-QN transfers as a
correction. Since the diagnostics read whichever tokeniser the audited
metric carries and IP is computed externally, the apparatus requires no
modification for such a study; only the metrics and their tokenisers need
substituting. The languages studied are five Indic abugidas; behaviour
under ideographic writing systems is untested, and to our knowledge no
corpus offers MQM-annotated evaluations of identical sources across
several non-Latin scripts, which is what the design requires. The
diagnostics themselves are computable on any script and any tokeniser, so
the apparatus would extend to Chinese, Japanese and Korean unchanged;
what is missing is the corpus, and we make no claim about how COMET-QN
would perform there. At
$1{,}400$ segments per language our corpus is smaller than those of
\citet{brahma-etal-2025-morphtok} and
\citet{kumar-etal-2017-morphological}; we claim only that our
dependent-vowel ordering agrees with theirs and so corroborates rather
than independently establishes it. The IPI zone boundaries and the
$\mathrm{SBI}\ge3.0$ flag are calibrated on this corpus, with the WMT24
controls providing a first external check but no more. COMET-QN assumes
comparable underlying quality across cells and would erase a genuine
quality difference between system pools, so it belongs within a single
dataset and domain; its reference distribution is a free parameter whose
influence on the pooled-agreement gain we have not measured. The
carried-over annotation design is supported by the encoder-dependence
ordering and the MAR counter-case (\S\ref{sec:restat},
\S\ref{sec:marathi}). We do not, however, measure transliteration
accuracy on this corpus. The published IndicXlit figures we quote in
\S\ref{sec:setup} are for the opposite direction, so residual
transliteration error in our own outputs is bounded by argument rather
than by direct measurement. A study collecting human ratings directly on
romanised output would
settle the question and remains future work, as does evaluation against
code-switched references, a complementary setting that our
fully-romanising design deliberately excludes.

\paragraph{Artefact licences:}
Every artefact used, namely
IndicMT~Eval~\citep{DBLP:conf/acl/BDNKKKD23}, the WMT24 evaluation
packages~\citep{kocmi-etal-2024-findings},
XLM-RoBERTa~\citep{DBLP:conf/acl/ConneauKGCWGGOZ20},
BLOOM-560m~\citep{lescao-etal-2023-bloom}, COMET,
BLEURT~\citep{DBLP:conf/acl/SellamDP20},
BERTScore~\citep{DBLP:conf/iclr/ZhangKWWA20} and
IndicXlit~\citep{madhani-etal-2023-aksharantar}, is released for research
use under a permissive or open-research licence. Our use is consistent
with those terms and we redistribute none of the underlying corpora.
Released code is MIT-licensed.

\section*{Ethics Statement}
\label{sec:ethics}
We foresee no ethical concerns arising from this work. The study measures
and corrects a distortion in automatic evaluation caused by tokenisation
artefacts affecting Indic scripts, and reports interpretable statistical
evidence together with an actionable procedure. All corpora and
benchmarks are publicly available and every released artefact is
analytical, with no foreseeable potential for harm. By exposing a
previously unmeasured distortion and supplying a correction for it, the
work is intended to benefit under-served Indic-language communities whose
systems are ranked by the affected metric.

\bibliography{custom}

\appendix
\setlength{\floatsep}{6pt}
\setlength{\textfloatsep}{8pt}
\setlength{\intextsep}{6pt}
\setlength{\abovecaptionskip}{3pt}
\setlength{\belowcaptionskip}{0pt}

\section{Statistical Detail and Sample Sizes}
\label{app:stats}
Table~\ref{tab:sample-sizes} gives per-language totals and MQM severity
counts. Five rows carrying non-numeric MQM entries are discarded, one
apiece in MAL and TAM and three in MAR, so per-language totals for
analyses needing a human score become $1{,}400$ for GUJ and HIN,
$1{,}399$ for TAM and MAL, and $1{,}397$ for MAR. No conclusion turns on
the choice of base: the ANOVA returns $F=2{,}080.8$ on the full corpus
and $F=2{,}080.2$ on the working set, so the $22.9\%$ variance result of
\S\ref{sec:obs} is unaffected and the discarded rows introduce no
detectable bias. Table~\ref{tab:fullcorr} gives agreement with MQM for all six
metrics in both conditions, the evidence behind the encoder-dependence
ordering of \S\ref{sec:restat}. Table~\ref{tab:axioms-stats} collects the
ANOVA, the SBI threshold proportions, the HIN--GUJ contrast and the
dependent-correlation tests. The Meng procedure belongs to the family of
the Williams test used for metric comparison at
WMT~\citep{graham-baldwin-2014-testing}; Steiger's
test~\citep{steiger-1980-tests} and a $5{,}000$-round paired permutation
test~\citep{deutsch-etal-2021-statistical} both agree with it. 

\begin{table}[tbp]
\centering\small
\setlength{\tabcolsep}{2.5pt}
\begin{tabular*}{\columnwidth}{@{\extracolsep{\fill}}lrrrrrrr@{}}
\toprule
\textbf{Lang.} & \textbf{N}
  & \textbf{Def} & \textbf{VL} & \textbf{L} & \textbf{M} & \textbf{H} & \textbf{VH} \\
\midrule
GUJ & $1{,}400$ & $248$ & $217$ & $326$ & $390$ & $184$ & $35$ \\
TAM & $1{,}400$ & $60$  & $124$ & $151$ & $323$ & $313$ & $429$ \\
MAL & $1{,}400$ & $372$ & $25$  & $162$ & $443$ & $334$ & $64$ \\
MAR & $1{,}400$ & $142$ & $69$  & $86$  & $184$ & $258$ & $661$ \\
HIN & $1{,}400$ & $399$ & $1$   & $7$   & $134$ & $324$ & $535$ \\
\bottomrule
\end{tabular*}
\caption{Per-language totals and MQM severity counts; buckets sum to
  $1{,}400$ in every language. \textbf{Def}\,=\,\texttt{Default}
  (error-free); VL/L/M/H/VH = Very Low, Low, Medium, High, Very High.
  Analyses needing a numeric human score use $1{,}400$ (GUJ, HIN),
  $1{,}399$ (TAM, MAL) and $1{,}397$ (MAR), giving the $6{,}995$-segment
  working set (\S\ref{sec:setup}).}
\label{tab:sample-sizes}
\end{table}

\begin{table*}[tbp]
\centering\small
\setlength{\tabcolsep}{6pt}
\begin{tabular*}{\textwidth}{@{\extracolsep{\fill}}lrrrrrr@{}}
\toprule
\textbf{Language}
  & \textbf{COMET} & \textbf{BLEURT} & \textbf{BERTScore}
  & \textbf{BLEU}  & \textbf{ChrF}   & \textbf{TER} \\
\midrule
GUJ$_\text{nat}$  & $\mathbf{+0.590\sig}$ & $+0.522\sig$ & $+0.438\sig$ & $+0.339\sig$ & $+0.401\sig$ & $-0.327\sig$ \\
GUJ$_\text{rom}$  & $+0.341\sig$ & $+0.339\sig$ & $+0.375\sig$ & $+0.340\sig$ & $+0.407\sig$ & $-0.333\sig$ \\
\addlinespace
TAM$_\text{nat}$  & $\mathbf{+0.629\sig}$ & $+0.563\sig$ & $+0.459\sig$ & $+0.320\sig$ & $-0.312^{\S}$ & $+0.410^{\S}$ \\
TAM$_\text{rom}$  & $+0.240\sig$ & $+0.377\sig$ & $+0.412\sig$ & $+0.286\sig$ & $+0.408^{\S}$ & $-0.308^{\S}$ \\
\addlinespace
MAL$_\text{nat}$  & $\mathbf{+0.651\sig}$ & $+0.596\sig$ & $+0.566\sig$ & $+0.470\sig$ & $+0.542\sig$ & $-0.451\sig$ \\
MAL$_\text{rom}$  & $+0.460\sig$ & $+0.443\sig$ & $+0.529\sig$ & $+0.409\sig$ & $+0.549\sig$ & $-0.466\sig$ \\
\addlinespace
MAR$_\text{nat}$  & $\mathbf{+0.487\sig}$ & $+0.443\sig$ & $+0.354\sig$ & $+0.261\sig$ & $+0.292\sig$ & $-0.255\sig$ \\
MAR$_\text{rom}$  & $+0.278\sig$ & $+0.286\sig$ & $+0.313\sig$ & $+0.256\sig$ & $+0.307\sig$ & $-0.258\sig$ \\
\addlinespace
HIN$_\text{nat}$  & $\mathbf{+0.422\sig}$ & $+0.381\sig$ & $+0.274\sig$ & $+0.202\sig$ & $+0.224\sig$ & $-0.236\sig$ \\
HIN$_\text{rom}$  & $+0.230\sig$ & $+0.294\sig$ & $+0.214\sig$ & $+0.204\sig$ & $+0.217\sig$ & $-0.213\sig$ \\
\bottomrule
\end{tabular*}
\caption{Spearman agreement between each metric and MQM on the working
  set ($N=1{,}400$ for GUJ and HIN, $1{,}399$ for TAM and MAL, $1{,}397$
  for MAR). Bold marks the strongest $|\rho|$ per language. COMET loses
  $|\Delta\rho|=0.19$ to $0.39$ under romanisation while the surface
  metrics stay flat. TER is negatively oriented.
  $^{\S}$TAM ChrF and TER reverse sign natively owing to
  character-$n$-gram fragmentation in the reference.}
\label{tab:fullcorr}
\end{table*}

\begin{table}[tbp]
\centering\small
\setlength{\tabcolsep}{4pt}
\begin{tabular*}{\columnwidth}{@{\extracolsep{\fill}}lrr@{}}
\toprule
\textbf{Quantity} & \textbf{Native} & \textbf{Romanised} \\
\midrule
$F$        & $2{,}080.8$  & $108.3$ \\
$p$        & $< 10^{-99}$ & $3.5 \times 10^{-25}$ \\
$\eta^{2}$ & $22.9\%$     & $1.5\%$ \\
\midrule
\multicolumn{3}{@{}l}{\textit{Segment-level $\mathrm{SBI}\ge 3.0$ proportions}} \\
GUJ & $62.1\%$ & $95.7\%$ \\
TAM & $34.9\%$ & $97.9\%$ \\
MAL & $30.9\%$ & $95.2\%$ \\
MAR & $28.2\%$ & $95.4\%$ \\
HIN & $\phantom{0}6.6\%$ & $98.3\%$ \\
\midrule
\multicolumn{3}{@{}l}{\textit{HIN--GUJ contrast (Welch's $t$)}} \\
Native    & \multicolumn{2}{c}{$13.95$ pts; $t=27.4\sig$, $p<10^{-99}$} \\
Romanised & \multicolumn{2}{c}{$\phantom{0}2.29$ pts; $t=6.97\sig$, $p=4.1\!\times\!10^{-12}$} \\
\midrule
\multicolumn{3}{@{}l}{\textit{Native-vs-romanised COMET--MQM decline
  (Meng $z$)}} \\
\multicolumn{3}{@{}l}{GUJ $10.5$; TAM $16.0$; MAL $9.8$; MAR $8.7$;
  HIN $10.1$} \\
\multicolumn{3}{@{}l}{\footnotesize Steiger $Z$ agrees to one decimal;
  $5{,}000$-round paired} \\
\multicolumn{3}{@{}l}{\footnotesize permutation: $0/5{,}000$ exceedances
  per language.} \\
\bottomrule
\end{tabular*}
\caption{ANOVA on COMET by script family ($N=7{,}000$; the ANOVA and the
  SBI proportions use metric-side quantities only, so no rows are
  dropped, and the same test on the $6{,}995$-segment working set gives
  $F=2{,}080.2$), the share of segments the $\mathrm{SBI}\ge3.0$ flag
  catches in each condition, the HIN--GUJ contrast, and the
  dependent-correlation tests
  \citep{meng-etal-1992-comparing,steiger-1980-tests,deutsch-etal-2021-statistical}.
  $\eta^{2}=\mathrm{SS}_{\text{between}}/\mathrm{SS}_{\text{total}}$.}
\label{tab:axioms-stats}
\end{table}

\section{Parity Profile and Diagnostic Definitions}
\label{app:langprof}
\begin{figure}[tbp]
  \centering
  \includegraphics[width=\columnwidth]{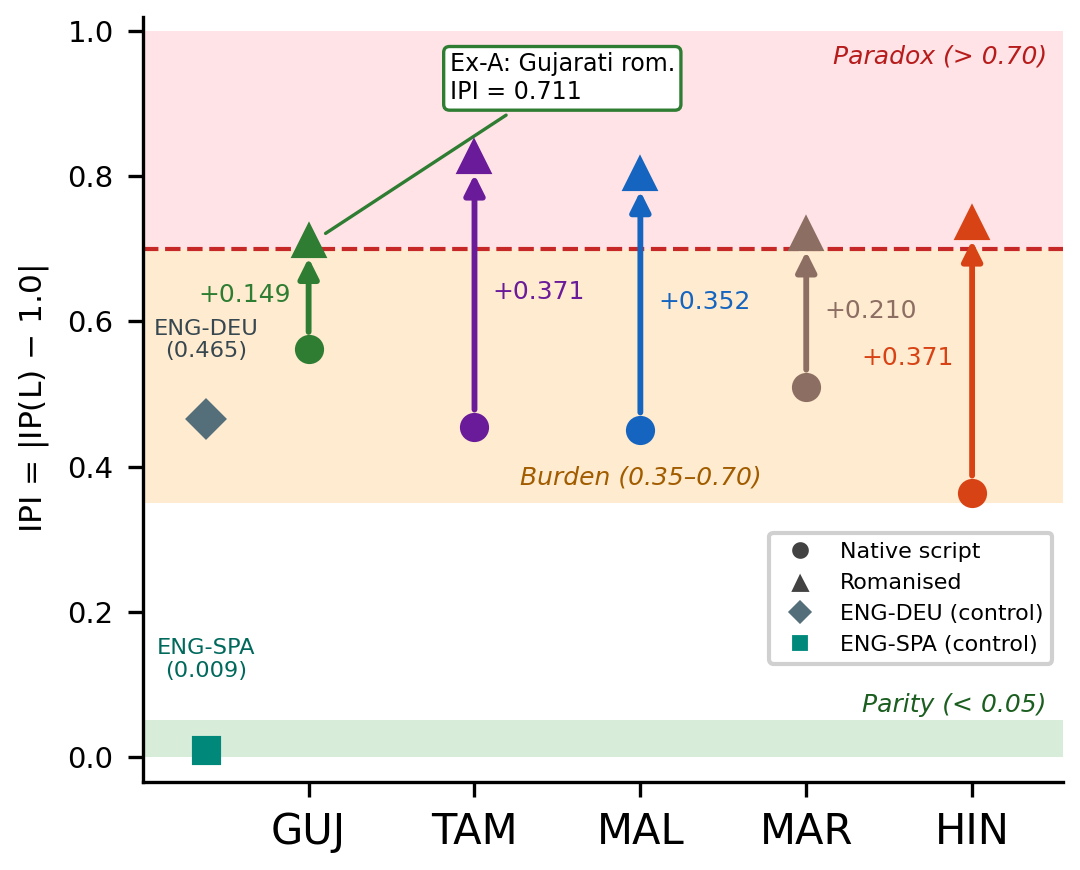}
  \caption{IPI zone transitions under romanisation. Bands: Parity
  ($<0.35$), Burden ($0.35$--$0.70$), Paradox ($>0.70$). Every Indic
  language crosses into Paradox; ENG-DEU (Burden) and ENG-SPA (Parity)
  are Latin-script reference points. The callout marks the romanised
  Malayalam example of Table~\ref{tab:motivating}(A), at
  $\mathrm{IPI}=0.803$ and COMET $=56.2$.}
  \label{fig:ipi_zones}
\end{figure}

Table~\ref{tab:diagnostics} defines the two prior parity measures and the
three diagnostics built from them. Table~\ref{tab:langprofile}
profiles all seven language-script settings, and
Figure~\ref{fig:ipi_zones} plots the IPI zone transitions.

\begin{table}[tbp]
\centering\small
\setlength{\tabcolsep}{4pt}
\resizebox{\columnwidth}{!}{%
\begin{tabular}{@{}l c p{3.6cm}@{}}
\toprule
\textbf{Metric} & \textbf{Ideal} & \textbf{Role} \\
\midrule
$\mathrm{TP}=|t(s_L)|/|t(s_E)|$ & $1.0$
  & Fragmentation cost; $>1$ over-fragmentation. \\
$\mathrm{IP}=\mathrm{NLL}(w_E)/\mathrm{NLL}(w_L)$ & $\approx 1.0$
  & Semantic density per token; $\ll 1$ collapse. \\
$\mathrm{SBI}=\mathbb{E}_i[\mathrm{TP}_i/\mathrm{IP}_i]$ & $1.0$
  & Burden per unit information. Flag $\ge 3.0$. \\
$\mathrm{IPI}=|\mathrm{IP}-1.0|$ & $0.0$
  & Distance from parity; Parity/Burden/Paradox zones. \\
$\mathrm{Tax}=\mathrm{LP}\times\mathrm{EP}$ & $1.0$
  & Romanisation overhead; LP, EP defined in \S\ref{sec:diag}. \\
\bottomrule
\end{tabular}}
\caption{The two prior parity measures (TP, IP) and the three diagnostics
  of \S\ref{sec:diag}. $t(\cdot)$ is XLM-RoBERTa's tokeniser
  and NLL the total negative log-likelihood under BLOOM-560m.
  \textbf{Ideal} is the value at English-equivalent parity.}
\label{tab:diagnostics}
\end{table}

\begin{table*}[tbp]
\centering\small
\setlength{\tabcolsep}{5pt}
\begin{tabular*}{\textwidth}{@{\extracolsep{\fill}}lllcccccccr@{}}
\toprule
\textbf{Lang.} & \textbf{Script} & \textbf{Family}
  & $\overline{\mathrm{IP}}_\mathrm{nat}$
  & $\overline{\mathrm{TP}}_\mathrm{nat}$
  & $\mathrm{SBI}_\mathrm{nat}$
  & $\mathrm{IPI}_\mathrm{nat}$
  & $\mathrm{IPI}_\mathrm{rom}$
  & \textsc{mattr}
  & DV/1k
  & IP--COMET $r$ \\
\midrule
SPA & Latin      & Romance   & $1.009$ & $1.090$ & $1.12$ & $0.009$ & n/a     & $0.581$ & n/a    & $-0.021\nsig$ \\
DEU & Latin      & Germanic  & $0.535$ & $1.133$ & $2.24$ & $0.465$ & n/a     & $0.634$ & n/a    & $-0.262\sig$ \\
GUJ & Gujarati   & IA        & $0.438$ & $1.388$ & $3.45$ & $0.562$ & $0.711$ & $0.752$ & $36.7$ & $-0.064\sig$ \\
TAM & Tamil      & Dravidian & $0.545$ & $1.319$ & $2.73$ & $0.455$ & $0.826$ & $0.862$ & $13.6$ & $-0.263\sig$ \\
MAL & Malayalam  & Dravidian & $0.549$ & $1.248$ & $2.62$ & $0.451$ & $0.803$ & $0.897$ & $42.5$ & $-0.266\sig$ \\
MAR & Devanagari & IA        & $0.490$ & $1.165$ & $2.61$ & $0.510$ & $0.720$ & $0.827$ & $24.5$ & $-0.216\sig$ \\
HIN & Devanagari & IA        & $0.636$ & $1.210$ & $2.03$ & $0.364$ & $0.735$ & $0.633$ & $18.3$ & $+0.203\sig$ \\
\bottomrule
\end{tabular*}
\caption{Parity profile for all seven settings. Indic values are from the
  Translation column of IndicMT~Eval~\citep{DBLP:conf/acl/BDNKKKD23}
  ($N=1{,}400$ per language); ENG-DEU and ENG-SPA are from
  WMT24~\citep{kocmi-etal-2024-findings} ($339$ sources pooled across MT
  systems, $6{,}006$ rows for ENG-DEU and $4{,}651$ for ENG-SPA).
  Controls appear first, then the fixed language order. DV/1k counts
  isolated dependent-vowel tokens per $1{,}000$ XLM-R tokens.
  IP--COMET $r$ is segment-level Pearson over all $1{,}400$ segments per
  language, requiring no human score. \textbf{n/a}: the Latin-script
  controls have no romanised condition. IA = Indo-Aryan. GUJ's IP--COMET
  correlation is weak but significant ($r=-0.064$, $p=0.017$).}
\label{tab:langprofile}
\end{table*}

\section{Computational Tax Decomposition}
\label{app:tax}
Table~\ref{tab:tax} splits the Computational Tax into its Length Penalty
(LP) and Entropy Penalty (EP) components for each language, with EP's
share of $\ln(\mathrm{Tax})$ in the final column.

\begin{table}[tbp]
\centering\small
\setlength{\tabcolsep}{5pt}
\begin{tabular*}{\columnwidth}{@{\extracolsep{\fill}}lcccc@{}}
\toprule
\textbf{Lang.} & \textbf{LP} & \textbf{EP} & \textbf{Tax} & \textbf{EP\%} \\
\midrule
GUJ & $1.15\times$ & $1.52\times$ & $1.75\times$ & $74.6\%$ \\
TAM & $1.78\times$ & $3.13\times$ & $5.57\times$ & $66.4\%$ \\
MAL & $1.48\times$ & $2.78\times$ & $4.12\times$ & $72.3\%$ \\
MAR & $1.41\times$ & $1.75\times$ & $2.47\times$ & $62.1\%$ \\
HIN & $1.41\times$ & $2.40\times$ & $3.37\times$ & $71.9\%$ \\
\bottomrule
\end{tabular*}
\caption{Length and Entropy Penalty components of the Computational Tax,
  with $\mathrm{EP\%}=\ln(\mathrm{EP})/\ln(\mathrm{Tax})$. The Entropy
  Penalty dominates in every language, so information lost per token,
  not tokens gained, is the principal cost of romanisation.}
\label{tab:tax}
\end{table}

\section{Dependent Vowels and Word Length}
\label{app:dv}
Vowels in Indic abugidas are written as diacritics attached to the
preceding consonant, and a well-matched tokeniser keeps each
consonant-vowel unit whole. XLM-RoBERTa's SentencePiece vocabulary,
induced from Latin-dominant text, frequently strands these diacritics as
tokens of their own carrying no meaning. MAL yields $42.5$ such isolated
tokens per $1{,}000$ XLM-R tokens and GUJ $36.7$
(Table~\ref{tab:langprofile}). TAM has the lowest rate at $13.6$ yet the
largest TP inflation under romanisation at $+77.9\%$. That dissociation
points to a second and independent mechanism. Agglutinative suffix chains
produce long word forms that the tokeniser must divide heavily, which
Table~\ref{tab:wltp} corroborates: TAM reaches $\rho=0.905\sig$ between
word count and token count in the romanised condition. MATTR is
vocabulary-free and hence independent of XLM-R, and it recovers the same
ordering (MAL $>$ TAM $>$ MAR $>$ GUJ $>$ HIN). That agreement points to
genuine morphological
diversity~\citep{hirak-etal-2026-assessing} rather than a
training-corpus artefact.

\begin{table}[tbp]
\centering\small
\setlength{\tabcolsep}{6pt}
\begin{tabular*}{\columnwidth}{@{\extracolsep{\fill}}lcc@{}}
\toprule
\textbf{Lang.} & \textbf{Native $\rho$} & \textbf{Romanised $\rho$} \\
\midrule
GUJ & $0.883\sig$ & $0.952\sig$ \\
TAM & $0.850\sig$ & $0.905\sig$ \\
MAL & $0.875\sig$ & $0.900\sig$ \\
MAR & $0.891\sig$ & $0.942\sig$ \\
HIN & $0.897\sig$ & $0.956\sig$ \\
\bottomrule
\end{tabular*}
\caption{Segment-level Spearman correlation between word count and XLM-R
  token count (all $p<10^{-99}$, $N=1{,}400$). Romanisation drives these
  toward $1.0$, the signature of single-character fallback.}
\label{tab:wltp}
\end{table}

\section{Latin-Script Controls}
\label{app:controls}
ENG-DEU and ENG-SPA from WMT24~\citep{kocmi-etal-2024-findings} serve as
Latin-script controls. SPA anchors the Parity zone at
$\mathrm{IPI}=0.009$, while DEU refutes the surface-script explanation:
written in the same alphabet as SPA, it nonetheless shows IP suppression
of the same magnitude as the Indic native range ($0.535$ against
$0.438$--$0.636$). The two diverge by a factor of twelve in their
IP--COMET correlation ($r=-0.262\sig$ for DEU against $r=-0.021\nsig$ for
SPA; $N=6{,}006$ and $4{,}651$). Byte premiums are nearly identical at
$1.25\times$ and $1.09\times$, which excludes character-encoding overhead
directly. German compounding drives the gap:
\textit{Erlassvereinfachungsvertr\"{a}ge}
(28 characters, ``decree simplification contracts'') costs $9$ XLM-R
tokens, whereas \textit{simult\'{a}neamente} (15 characters,
``simultaneously'') costs $4$. Only $14.6\%$ of DEU word types receive a
single-token encoding against $22.1\%$ for SPA, and long DEU words need
$32\%$ more tokens than long SPA words. DEU's MATTR of $0.634$ exceeds
SPA's $0.581$, consistent with the compounding account.
Table~\ref{tab:controls} sets both controls beside the Indic native
range.

\paragraph{A reversal in SBI:}
Across the Indic languages, script-induced SBI elevation is independent
of quality: fragmentation applies as a structural offset irrespective of
how good a translation is. In DEU the relationship inverts. Erroneous MT
output there tends toward paraphrase and syntactic
simplification~\citep{DBLP:conf/wmt/ZouharBBN24}, producing shorter and
commoner word sequences and hence \emph{less} fragmentation. Error-free
DEU segments average $\mathrm{SBI}=2.349$ while segments with major
errors average $2.129$ (Kruskal--Wallis $H=135.0\sig$,
$p=5.0\times10^{-30}$), the reverse of the romanised-Indic direction. SBI
nonetheless remains a valid zone diagnostic in both cases, placing DEU
in Burden at $2.24$ and SPA in Parity at $1.12$.

\begin{table*}[tbp]
\centering\small
\setlength{\tabcolsep}{6pt}
\begin{tabular*}{\textwidth}{@{\extracolsep{\fill}}lcccccc@{}}
\toprule
\textbf{Setting}
  & $\overline{\mathrm{IP}}$
  & \textbf{SBI}
  & \textbf{MATTR}
  & \textbf{Byte\,pr.}
  & \textbf{1-tok\%}
  & \textbf{IP--COMET $r$} \\
\midrule
ENG-SPA (SPA) & $1.009$ & $1.12$ & $0.581$ & $1.09\times$ & $22.1\%$ & $-0.021\nsig$ \\
ENG-DEU (DEU) & $0.535$ & $2.24$ & $0.634$ & $1.25\times$ & $14.6\%$ & $-0.262\sig$ \\
\midrule
Indic nat. & $0.438$--$0.636$ & $2.03$--$3.45$ & $0.633$--$0.897$
   & $2.45$--$5.33\times$ & $8.7$--$30.3\%$ & varies \\
\bottomrule
\end{tabular*}
\caption{Latin-script controls beside the Indic native range.
  \textbf{1-tok\%} is the share of unique word types receiving a
  single-token XLM-R encoding. Zones: ENG-SPA Parity; ENG-DEU Burden;
  Indic native Burden; Indic romanised Paradox.}
\label{tab:controls}
\end{table*}

\section{Vocabulary-Free Corroboration}
\label{app:corp}
Table~\ref{tab:corroboration} reports MATTR and Byte Premium across all
seven settings. Both are computed without reference to any model
vocabulary, so they check the TP and IP findings independently of XLM-R
and of UTF-8 encoding.

\begin{table}[tbp]
\centering\small
\setlength{\tabcolsep}{4pt}
\resizebox{\columnwidth}{!}{%
\begin{tabular}{@{}llccc@{}}
\toprule
\textbf{Lang.} & \textbf{Script}
  & \textbf{MATTR\textsubscript{nat}}
  & \textbf{MATTR\textsubscript{rom}}
  & \textbf{Byte Prem.} \\
\midrule
GUJ & Gujarati   & $0.752$ & $0.665$\,$\downarrow$ & $3.03$ \\
TAM & Tamil      & $0.862$ & $0.733$\,$\downarrow$ & $5.30$ \\
MAL & Malayalam  & $0.897$ & $0.768$\,$\downarrow$ & $5.33$ \\
MAR & Devanagari & $0.827$ & $0.721$\,$\downarrow$ & $3.64$ \\
HIN & Devanagari & $0.633$ & $0.601$\,$\downarrow$ & $2.45$ \\
\midrule
SPA & Latin (Romance)  & $0.581$ & n/a  & $1.09$ \\
DEU & Latin (Germanic) & $0.634$ & n/a  & $1.25$ \\
\bottomrule
\end{tabular}}
\caption{MATTR and Byte Premium across all seven settings as
  vocabulary-free and encoding-independent checks on the TP and IP
  findings. MATTR uses $w=500$ with whitespace
  tokenisation~\citep{covington-mcfall-2010-mattr}; Byte Premium is the
  target-to-English ratio of UTF-8 bytes per
  word~\citep{arnett-etal-2024-bit,arnett-bergen-2025-byte}. $\downarrow$
  marks a fall under romanisation; \textbf{n/a}: no romanised condition
  exists for the Latin-script controls.}
\label{tab:corroboration}
\end{table}

\section{Axiomatic Motivation and Threshold Calibration}
\label{app:axioms}
\paragraph{Why SBI takes a ratio form:}
A measure of representational burden ought to (a)~equal $1.0$ when
$\mathrm{TP}=\mathrm{IP}=1.0$; (b)~be unaffected by constant rescaling of
either input; (c)~grow without bound as fragmentation rises or density
collapses; and (d)~identify the GUJ case, where the smallest TP inflation
under romanisation of the five ($+15.2\%$) combines with the lowest
native information parity ($\overline{\mathrm{IP}}_\mathrm{nat}=0.438$)
to produce the greatest overall burden. The additive form
$\mathrm{TP}-\mathrm{IP}$ violates (a) and (d). The ratio
$\mathrm{TP}/\mathrm{IP}$ is the simplest function meeting all four.

\paragraph{Why Tax multiplies:}
The two charges are incurred at once: a segment requires more tokens (LP)
and each conveys less (EP). Because neither can be exchanged for the
other, their joint effect on inference cost compounds rather than
accumulates. TAM at $\mathrm{LP}=1.78$ and $\mathrm{EP}=3.13$ is a
different situation from $\mathrm{LP}=3.13$ and $\mathrm{EP}=1.78$
despite the identical sum.

\paragraph{Calibrating the IPI boundaries:}
The Paradox boundary at $0.70$ is set empirically. Romanised agreement
values are $\{0.230, 0.240, 0.278, 0.341, 0.460\}$ across the five
languages, four of which fall below $\rho=0.40$; MAL at $0.460$ is the
exception, the single romanised setting above $\rho=0.40$ in the Paradox
zone, and it still shows an absolute decline of $0.191$ from its native
$0.651$, with shared variance halved from $0.42$ to $0.21$. The Burden
upper bound is the largest IPI compatible with $\rho\ge0.40$ across all
native-script settings, and $0.70$ is a round value inside the empirical
gap separating the highest native setting (GUJ at $0.562$) from the
lowest romanised one (GUJ at $0.711$). Consequently all twelve
classifications are insensitive to where in that gap the boundary is
drawn, which is what makes the zone labels of \S\ref{sec:diag} robust to
the choice. The Parity/Burden boundary is set at $0.35$, below the lowest
Burden setting (HIN native at $0.364$) and well above SPA's $0.009$; no
setting in our data falls in the $0.05$--$0.35$ interval, which is
therefore left inside Parity by construction.

\section{COMET-QN Procedure and Reproducibility}
\label{app:qnproc}
\paragraph{Working set:}
Numeric fields are coerced and the five rows carrying non-numeric MQM
entries are dropped, leaving $6{,}995$ records, each holding
native-script and romanised COMET, a human MQM score, and TP and IP for
both conditions. Analyses that use no human score retain all $7{,}000$
rows (\S\ref{sec:setup}).

\paragraph{Quantile renormalisation (Eq.~\ref{eq:qn}):}
Fix the reference $R$ as the pooled native-script COMET distribution over
all languages, sorted. Within each cell, replace every score by its
fractional rank $\mathrm{rank}(s_i)/(n+1)$, then map that fraction
through the empirical quantile function $Q_R$. Both steps are monotone,
hence rank-preserving within the cell. Table~\ref{tab:qnexample} works
five romanised Gujarati segments through the map.

\begin{table}[tbp]
\centering\small
\setlength{\tabcolsep}{6pt}
\begin{tabular*}{\columnwidth}{@{\extracolsep{\fill}}rrrr@{}}
\toprule
\textbf{Raw} & \textbf{Rank} & \textbf{Position} & \textbf{Mapped} \\
\midrule
$71.07$ & $1$ & $0.167$ & $68.21$ \\
$76.41$ & $2$ & $0.333$ & $77.27$ \\
$78.14$ & $3$ & $0.500$ & $83.14$ \\
$78.24$ & $4$ & $0.667$ & $87.58$ \\
$80.33$ & $5$ & $0.833$ & $91.48$ \\
\bottomrule
\end{tabular*}
\caption{Eq.~\ref{eq:qn} applied to five romanised GUJ segments.
  Position is $\mathrm{rank}/(n{+}1)$; Mapped is the value at that
  position of the pooled native-script reference. Ranks are carried
  through unchanged, which is why the operation cannot disturb any
  within-cell ordering.}
\label{tab:qnexample}
\end{table}

\paragraph{Sensitivity regressor:}
Gradient-boosted trees with $300$ estimators, depth $3$, learning rate
$0.05$ and seed $0$. Features are $\mathrm{COMET}_\mathrm{rom}$, the four
parity values $\mathrm{TP}_\mathrm{nat}$, $\mathrm{TP}_\mathrm{rom}$,
$\mathrm{IP}_\mathrm{nat}$ and $\mathrm{IP}_\mathrm{rom}$, and the two
differences
$\Delta\mathrm{TP}=\mathrm{TP}_\mathrm{nat}-\mathrm{TP}_\mathrm{rom}$ and
$\Delta\mathrm{IP}=\mathrm{IP}_\mathrm{nat}-\mathrm{IP}_\mathrm{rom}$.
The target is $\mathrm{COMET}_\mathrm{nat}$.

\Needspace*{7\baselineskip}
\section{The Reporting Procedure on One Setting}
\label{app:worked}
This appendix applies \S\ref{sec:rec} end to end to romanised Gujarati,
using only numbers already reported above.

\textbf{Step 1, name the tokeniser.} The metric is
\texttt{wmt22-comet-da}, which carries the XLM-R SentencePiece
vocabulary (\S\ref{sec:setup}). It is inside the profiled family, so the
remaining steps apply unchanged.

\textbf{Step 2, compute IPI.} For romanised Gujarati,
$\mathrm{IPI}=0.711$ (Table~\ref{tab:langprofile}). No human ratings are
involved.

\textbf{Step 3, read the zone.} $0.711$ exceeds $0.70$, so the setting is
in Paradox. Steps 3 and 4 do not apply; step 5 does.

\textbf{Step 5, assemble the report.} Publish COMET-QN as the headline
figure, with the raw mean COMET of $70.61$ beside it, and the three
diagnostics: $\mathrm{SBI}_\mathrm{nat}=3.45$,
$\mathrm{IPI}_\mathrm{rom}=0.711$, and $\mathrm{Tax}=1.75\times$. Do not
compare the raw score against a native-script score from another
language. The regressor gate then applies. Gujarati's mean cell gap is
negative, native minus romanised giving $-15.09$ points, the same sign as
TAM, MAL and HIN, so a corrector fitted on those languages transfers and
recovers $38.6\%$ of the lost sensitivity (Table~\ref{tab:lolo}). Had the
sign matched MAR instead, the gate would have withheld the regressor and
left COMET-QN alone.

\textbf{Step 6, do not romanise.} Romanising Gujarati to equalise scripts
would cost $1.75\times$ the encoder compute and drop agreement with
annotators from $\rho=0.590$ to $0.341$.

A reader of that report can see the corrected score, the uncorrected
score, how heavily the tokeniser taxed the language, and how much
distortion the correction did not remove.

\paragraph{Protocols:}
Table~\ref{tab:impossibility} uses a 50/50 within-language split at seed
$42$; Figure~\ref{fig:qn} uses the whole working set;
Table~\ref{tab:lolo} uses leave-one-language-out. Correlations are
segment-level Spearman unless labelled Pearson, and ``pooled''
concatenates cells before correlating. All values are regenerated by the
released analysis script.
\balance
\end{document}